\documentclass{article}

\usepackage{arxiv}
\usepackage{amsmath}

\usepackage[utf8]{inputenc} 
\usepackage[T1]{fontenc}    
\usepackage{hyperref}       
\usepackage{url}            
\usepackage{booktabs}       
\usepackage{amsfonts}       
\usepackage{nicefrac}       
\usepackage{microtype}      
\usepackage{cleveref}       
\usepackage{lipsum}         
\usepackage{graphicx}
\usepackage[square,sort,comma,numbers]{natbib}
\usepackage{doi}
\usepackage{textcomp}
\usepackage{booktabs}
\usepackage{subcaption}
\usepackage{xcolor}

\title{ConKeD: Multiview contrastive descriptor learning for keypoint-based retinal image registration}

\newif\ifuniqueAffiliation
\uniqueAffiliationtrue

\ifuniqueAffiliation 

\usepackage{authblk}

\newbox{\orcid}\sbox{\orcid}{\includegraphics[scale=0.06]{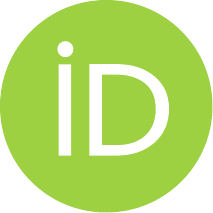}} 
\author[1,2]{%
	\href{https://orcid.org/0000-0001-7824-8098}{\usebox{\orcid}\hspace{1mm}David Rivas-Villar\thanks{\texttt{Corresponding author: david.rivas.villar@udc.es}}}%
}
\author[1,2]{%
	\href{https://orcid.org/0000-0002-9080-9836}{\usebox{\orcid}\hspace{1mm}Álvaro S. Hervella\thanks{\texttt{a.suarezh@udc.es}}}%
}
\author[1,2]{%
	\href{https://orcid.org/0000-0003-4407-9091}{\usebox{\orcid}\hspace{1mm}José Rouco \thanks{\texttt{jrouco@udc.es}}}%
}
\author[1,2]{%
	\href{https://orcid.org/0000-0002-0125-3064}{\usebox{\orcid}\hspace{1mm}Jorge Novo\thanks{\texttt{jnovo@udc.es}}}%
}
\affil[1]{Grupo VARPA, Instituto de Investigacion Biomédica de A Coru\~na (INIBIC), Universidade da Coru\~na, 15006 A Coru\~na, Spain}
\affil[2]{Departamento de Ciencias de la Computación y Tecnologías de la Información, Universidade da Coru\~na, A Coruña, 15071, A Coruña, Spain}
\fi

\renewcommand{\shorttitle}{ConKeD}

\hypersetup{
pdftitle={ConKeD: Multiview contrastive descriptor learning for keypoint-based retinal image registration},
pdfsubject={cs.CV},
pdfauthor={David~Rivas-Villar, Álvaro S.~Hervella, José~Rouco, Jorge~Novo},
pdfkeywords={Self Supervised Learning, Feature-based Registration, Image Registration, Retinal Image Registration, Medical Imaging},
}

\begin{document}
\maketitle

\begin{abstract}
Retinal image registration is of utmost importance due to its wide applications in medical practice. In this context, we propose ConKeD, a novel deep learning approach to learn descriptors for retinal image registration. In contrast to current registration methods, our approach employs a novel multi-positive multi-negative contrastive learning strategy that enables the utilization of additional information from the available training samples. This makes it possible to learn high quality descriptors from limited training data. To train and evaluate ConKeD, we combine these descriptors with domain-specific keypoints, particularly blood vessel bifurcations and crossovers, that are detected using a deep neural network. Our experimental results demonstrate the benefits of the novel multi-positive multi-negative strategy, as it outperforms the widely used triplet loss technique (single-positive and single-negative) as well as the single-positive multi-negative alternative. Additionally, the combination of ConKeD with the domain-specific keypoints produces comparable results to the state-of-the-art methods for retinal image registration, while offering important advantages such as avoiding pre-processing, utilizing fewer training samples, and requiring fewer detected keypoints, among others. Therefore, ConKeD shows a promising potential towards facilitating the development and application of deep learning-based methods for retinal image registration.

\end{abstract}


\section{Introduction}
\label{sec:introduction}
Image registration is the process of spatially aligning a pair of images. In this process, one image is used as the reference, the fixed image, while the second image, the moving image, is spatially transformed to match the first one. These images are usually from the same subject and they differ in their point of view or in the instant they were captured in. \ 
In medicine, image registration is particularly important due to its multiple applications on clinical practice \cite{survey}. Image registration enables the simultaneous analysis of multiple images which, in turn, allows clinicians to draw more informed conclusions \cite{book_mir}. Additionally, medical image registration enables longitudinal studies which are useful to monitor disease progression or remission \cite{Narasimha}.

Retinal Image Registration (RIR) is a highly relevant task as the eyes are the only organs in the human body that allow non-invasive \textit{in vivo} observation of the blood vessels and neuronal tissue \cite{forrester2020eye}. In particular, there is a high interest in the registration of color fundus (CF) images due to the widespread use of this image modality in clinical practice \cite{RIR}. CF images are cost effective due to the relatively low price of CF cameras combined with their efficacy in the diagnosis of numerous diseases \cite{costeffec,Kanski}. However, CF images also present particular characteristics that complicate the registration. For instance, their photographic nature can cause several imperfections due to incorrect placement of the device, incorrect capture by the clinician or movements by the patient. Some of these imperfections are color or illuminance variation, changes in focus, halos, etc. Moreover, certain diseases can drastically alter the retinal appearance. These morphological changes complicate the registration process in the presence of pathology. The combination of all of these characteristics makes the CF registration process a challenging task.

In broad terms, automatic registration approaches can be classified according to their algorithmic basis as either classical approaches or deep learning approaches. Furthermore, registration methods can also be divided into three main groups: Feature Based Registration (FBR), which uses keypoints and descriptors; Intensity Based Registration (IBR), which uses the intensity values of the image; and Direct Parameter Regression (DPR), which predicts a deformation field or matrix directly from the input images \cite{rivas2, Rivas-Villar:23}.

Classical registration methods are still widely used nowadays. However, novel deep learning approaches have several advantages that usually make them more desirable over the classical methods. For instance, deep learning methods can be trained end-to-end from raw data to the expected result. This eliminates the need for ad-hoc feature engineering which allows deep learning methods to be more flexible and adaptable to changes in the input data. 

FBR methods use keypoints to drive the registration process. These keypoints are distinctive spatial locations that can be detected in the different images to register. The keypoints are matched among the images and used to compute the geometric transformation that aligns them. FBR methods are inherently explainable as the keypoints used to calculate the transformation are known and can be visualized. Commonly, to aid in the matching process, point descriptors are also computed. These descriptors are unique feature representations that characterize each keypoint and facilitate their distinction. FBR methods are divided into methods that use generic keypoints (valid for any domain) and domain-specific keypoints (generally more accurate but only usable in a single domain). Generic methods are potentially less accurate and more computationally expensive due to the lower detection specificity that makes necessary the detection and processing of a higher number of keypoints.

IBR approaches work by iteratively maximizing a similarity metric between the intensity values of the images to be registered across transformation parameter space. There are multiple suitable similarity metrics \cite{Pluim,Balakrishnan}, including deep learning-based ones \cite{cheng18}.


DPR methods allow direct transformation prediction or deformation field prediction which, in turn, allows image alignment \cite{Haskins,voxel}. To achieve this, a deep neural network is trained such that its output is used to transform the moving image. In this case, the training objective is to maximize the similarity between the transformed moving image and the fixed image or, equivalently, between two simplified representations of these images (such as e.g. blood vessel segmentation maps for CF images \cite{Benvenuto}). The key difference between IBR and DPR is how they use similarity metrics. IBR explicitly maximizes a similarity metric while DPR uses the similarity metric to guide the learning of neural network. This network learns to create suitable transformations directly from the input data, only relying on the similarity metric as guidance during its training step.

Finally, there are hybrid methods joining several of these registration paradigms. In this case, FBR methods usually provide an initial alignment which is then refined by other IBR or DPR methods that are capable of deformable registration thus improving the results \cite{alvaro,Rv-v}. 
However, while these approaches might be successful in terms of results, they are not desirable since they increase the complexity and the computational cost and may introduce extra hyperparameters.

Currently, most state-of-the-art medical registration methodologies are based on IBR or DPR methods \cite{voxel,dlir,Haonan}. However, RIR tasks, and particularly CF images, have certain characteristics that limit the applicability of these methods. In particular, the patterns that are useful for the registration of CF images (i.e. blood vessel bifurcations and crossovers, optic disc, etc.) are relatively small and are scattered over a homogeneous background. Furthermore, CF images can present large displacements and low degree of overlapping between the images in each pair. Additionally, there can be multiple intra-pair differences produced by the photographic nature of the capture process (i.e. light halos, blurriness due to suboptimal focus, etc.). These changes are specifically detrimental to IBR methods as creating a robust similarity metric is challenging under these conditions. Similarly, as the transformations on the retina are almost always rigid, deformable registration methods like DPR using deformation fields are not suitable as they pose the risk of overfitting these fields creating unrealistic transformations. Therefore, commonly used methods in other medical image modalities such as  \cite{voxel,dlir} are unsuitable for color fundus registration.

Traditionally, the CF registration field has been dominated by classical FBR methods \cite{rempe,votus}. Some of these methods, such as REMPE and VOTUS, still obtain the best results in most of the categories of the well-known FIRE benchmark dataset \cite{rempe,votus}. REMPE \cite{rempe} combines generic points (SIFT) and domain-specific keypoints (blood vessel bifurcations). Then, using RANSAC (Random Sample Consensus) \cite{ransac} and Particle Swarm Optimization, the transformation is computed according to an ellipsoidal model of the eye that has been created specifically for this task. Differently, VOTUS \cite{votus} creates graphs for the arterio-venous tree. Then, the graphs from the corresponding images are matched using a novel algorithm (Vessel Optimal Transform) that relies on classical image features. The transformation matrix is computed using DeSAC (Deterministic Sample And Consensus).

Recently, some novel FBR deep learning based methods have been able to compete with and even outperform the classical approaches in some cases \cite{rivas,rivas2,eccv20}, whereas previous DPR deep learning methods could not \cite{zou}. Some of these methods are based on domain-specific keypoints \cite{rivas,eccv20} while others employ generic keypoint detectors \cite{rivas2}. 

The work of Rivas-Villar et. al. \cite{rivas} was the first FBR deep learning method for CF image registration. It uses a CNN to detect blood vessel crossovers and bifurcations which are then used to infer a transformation using RANSAC without the need for descriptor computation. However, the lack of descriptors limits the complexity of the transformation as calculating all the possible combinations of keypoints with RANSAC by brute-force increases in computational cost with the degrees of freedom in the geometric transformation. SuperRetina \cite{eccv20} adapted SuperPoint \cite{superpoint}, a natural image registration method, to RIR. They propose to train this network using keypoints labeled by an expert manually or by a classical approach, known as PBO \cite{pbo}. To perform the matching of keypoints, they use descriptors that are learnt using a triplet loss approach. Their method requires extensive ad-hoc image pre-processing, firstly taking only the green channel from the RGB images, normalizing it, applying CLAHE contrast enhancement \cite{CLAHE}, and finally using gamma adjustment. These pre-processing steps simplify the task but require multiple parameters that may need to be tuned on a per-dataset or per-device basis. After this, the keypoints and descriptors are jointly trained starting from a set of labeled keypoints which are progressively expanded using the more confident detections of the network outside of the ground truth. In order to complete the registration pipeline, SuperRetina uses a double inference step. This means that, the registration is firstly done normally and, after transforming the moving image, the process is repeated again to improve the results at the expense of more computation cost. Similarly, in \cite{rivas2} R2D2 \cite{r2d2} is adapted for the retinal image registration task. R2D2 is a natural image registration method capable of jointly learning keypoints and descriptors without ground truth data, making this method fully unsupervised. Unlike SuperRetina, this method does not require pre-processing or double inference.
However, it obtains worse results. In this method the descriptors are learnt using a multi-negative loss, known as AP loss.

Overall, using domain-specific keypoints, such as blood vessel crossovers and bifurcations, involves the use of supervised learning and manually labeled data \cite{rivas,eccv20}, whereas methods which use generic keypoint detectors do not require labeled data \cite{rivas2}. However, due to the lack of precise and highly distinctive keypoints, generic detectors usually carry an increased computational cost. 
In terms of descriptor learning, deep learning methods use contrastive learning, a technique based on comparing samples. For a given keypoint, which is used as reference (anchor keypoint), there are positive samples (the same point in another view of the same image) and negative samples (any other point in the same or another view). The main difference among methods is how they use the different samples and how many samples they are able to process on each contrastive learning step. 
Currently, in natural image registration as well as in RIR, most approaches use triplet loss \cite{eccv20, superpoint} which exploits a single positive and negative sample per anchor. 
Moreover, in RIR, there are some approaches that employ multi-negative single-positive methods \cite{rivas2} or do not use descriptors at all \cite{rivas}. 
In this regard, none of the state-of-the-art methods use novel multi-positive multi-negative approaches for image registration, neither in natural images nor in medical images. Multiple positive contrastive learning has been successfully used in other tasks, such as image classification, where it has demonstrated improved performance over just multi-negative approaches \cite{supcon}. These methods employ multiviewed batches to create multiple positives for each anchor, which in turn, allows to leverage more information from the input images. However, no multi-positive multi-negative framework has been developed for keypoint-based registration despite the success and benefits that this self-supervised approach demonstrated in other areas \cite{moco, simclr}.

In this work, we propose ConKeD, \textbf{Con}trastive \textbf{Ke}ypoint \textbf{D}escriptors, a novel deep learning method to learn descriptors for domain-specific keypoint-based RIR. In contrast to previous methods, we use a multi-positive multi-negative contrastive learning approach that allows to leverage additional information from the input images, hence improving the learning of descriptors. Our multi-positive framework represents the first approach of this kind for image registration. In that regard, we propose a complete methodology for the registration of CF images, using vessel crossovers and bifurcations as keypoints together with the descriptors learned using ConKeD. These keypoints are highly specific and unique which helps reducing the number of required detections to perform the registration. 
The main contributions of the paper are the following:
\begin{itemize}
    \item We propose the first multi-positive multi-negative contrastive learning approach for keypoint-based image registration.

    \item We propose a complete RIR deep learning methodology that takes advantage of domain-specific keypoints and our novel descriptors.
    
    \item We provide extensive experiments in the public FIRE dataset comparing ConKeD against established baselines and analyzing different factors of the method.

    \item Our experimentation reveals that the novel multi-positive multi-negative approach improves the performance of the commonly used triplet loss as well as multi-negative approaches.

    \item Our proposal achieves results comparable to those in the state-of-the-art while avoiding image pre-processing, requiring less training samples, and also requiring less detected keypoints.

\end{itemize}

\section{Materials and Methods}

\subsection{Methodology}

\begin{figure*}
    \centering
    \includegraphics[width=\textwidth]{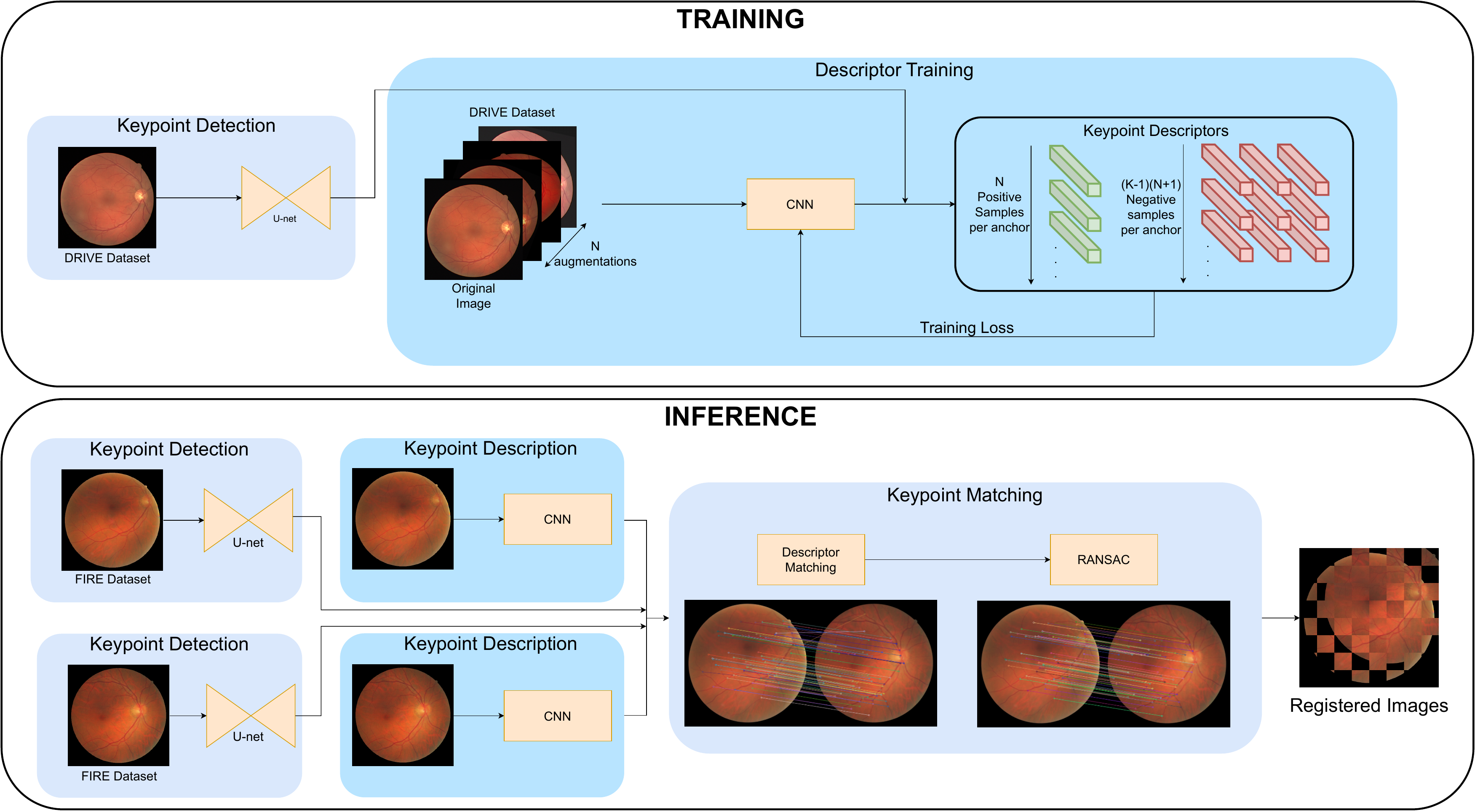}
    \caption{Overview of the proposed method for both training and inference.}
    \label{fig:over}
\end{figure*}

The proposed methodology for CF registration is described in Fig. \ref{fig:over}. This methodology is based on ConKeD, our novel framework for learning descriptors for keypoint-based image registration. These keypoints are detected using a state-of-the-art neural network \cite{rivas}. Meanwhile, another network is used to compute the descriptors of these keypoints. The descriptor network is trained with a multiviewed batch containing one original image and $N$ views or augmentations of the original image. Therefore, each keypoint (i.e. anchor) will have $N$ positive samples while all the rest of detected keypoints act as negatives, including the points in the own image. These samples coupled with a suitable loss allow the keypoint descriptor to be trained, learning to match the positive samples and differentiate the negatives.
At inference time, the keypoints are detected, described, matched and, finally, RANSAC is used to infer a projective transform (homography) to align both images.

\subsubsection{Keypoint detection}

The first step of the proposed methodology is to detect blood vessel crossovers and bifurcations. These keypoints are highly specific and distinctive, allowing for accurate matching. However, the number of these keypoints in each image is unknown a priori and can even change between images of the same individual (i.e. in the same pair). This is due to the difference in viewpoint as well as pathology progression, which can occlude or banish some keypoints.

In order to detect these keypoints, we train a CNN to generate heatmaps depicting all the possible keypoints in the input image \cite{alvaro_cmpb}. Using heatmaps instead of binary maps is beneficial due to the heavy imbalance between the positive class (the keypoint pixels) and the negative class (the background pixels)\cite{alvaro_cmpb}. The produced heatmaps have maximum values at the location of the detected keypoints and progressively decreasing values in the neighboring pixels. The heatmaps increase the information available to the network during training, improving the learning process \cite{alvaro_cmpb}.

To distinguish between crossovers and bifurcations, two separate heatmaps are generated, one for each type of keypoint. Moreover, to incentivize the network to detect all the keypoints even if their type could not be inferred, a third output heatmap containing both types of keypoints is also generated. The network is trained using the mean squared error (MSE) between the predicted heatmaps and the target heatmaps as the loss function. In order to create the target heatmaps, the original binary ground truth is convolved using a Gaussian kernel.

Finally, during inference, in order to extract the keypoints, we compute the heatmaps using this network. From the heatmaps we can obtain the discrete locations of the keypoints by using a local maxima filter and an intensity threshold \cite{alvaro_cmpb}. 
 
\subsubsection{Keypoint description}

\begin{figure*}
    \centering
    \includegraphics[width=\textwidth]{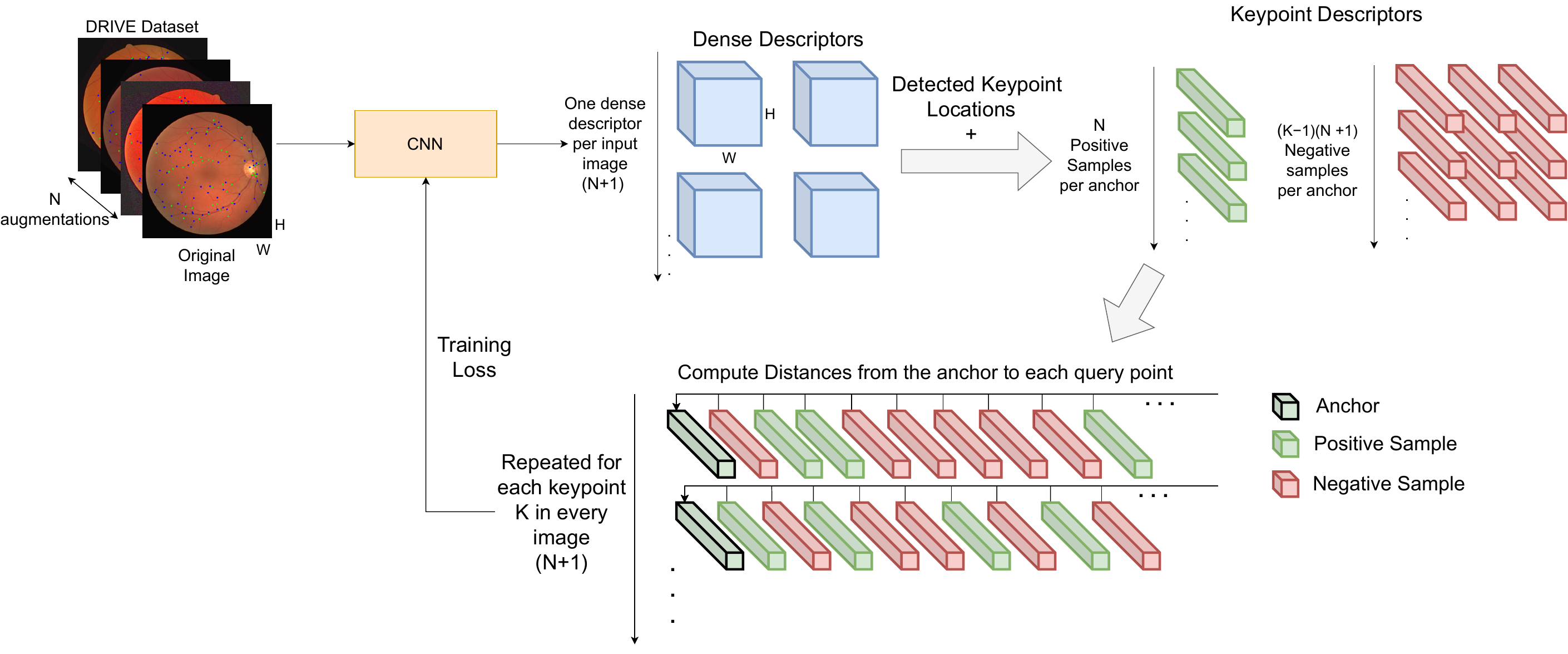}
    \caption{ConKeD methodology for description learning.}
    \label{fig:descs}
\end{figure*}

The second step is to generate the descriptors for the detected keypoints. For this step, we propose a novel framework, ConKeD, which is schematically described in Fig. \ref{fig:descs}. 

ConKeD follows a novel multi-positive multi-negative contrastive learning approach in order to learn descriptors for image registration. Typically, deep learning-based FBR methods learn descriptors using triplet loss approaches. These approaches compare each given point against both a positive and a negative sample. Some recent methods also explored multi-negative approaches which, instead of selecting a particular negative sample, use of all of them. In contrast, our method leverages multiple samples for both positive and negative examples. This avoids the need for explicit mining and increases the amount of contrastive comparisons, facilitating training.

The fundamental idea behind our proposal is to compare the representations (i.e. descriptors) of different keypoints. In particular, the key concept is to pull together the anchor (a particular point used as reference) and all the positive samples in the descriptor space while pushing apart the anchor from all the different negative samples.  

In order to process multiple positive and multiple negative samples during training, we create a multiviewed batch for each input image during training. This batch includes an image from the dataset, serving as base, and $N$ views or image augmentations. Thus, our multiviewed batch is of size $1+N$.
The image augmentations (views) are generated from the base image using  spatial and color transformations. Therefore, all of the positive samples are augmentations of the anchor keypoint while negative samples are the rest of the detected keypoints, belonging to any image in the batch. In any case, it should be noted that ConKeD does not require multiple images from the same individual and, therefore, it could be trained using unlabeled data.

Additionally, in order to maximize the number of comparisons, our novel approach is trained with the whole images (as opposed to patches). For each image, the CNN produces a dense block of descriptors, each descriptor corresponding to one pixel of the input image. Therefore, for each keypoint location detected in the previous step, it is possible to directly select its specific descriptors. Considering that we have an average of $K$ keypoints per image and a multiviewed batch size $1+N$, there are a total of $K(1+N)$ keypoint samples per batch. In particular, each one of the $K$ keypoints will have $N$ positive samples, and $(K-1)(N+1)$ negative samples. As we do not perform any explicit negative point mining, for each keypoint, every other keypoint that is not a positive sample is considered a negative sample, including the ones on the same image where this keypoint was detected. However, in this case, all of the negative samples can be considered hard-negatives. This is because they are all from the set of detected keypoints and, therefore, they are either vessel crossovers or bifurcations (i.e. closely related points). Finally, given the multiple positives and negatives for each anchor point, all the possible pairings are created and all the possible comparisons are computed between the descriptors.

In order to compare the descriptors, we use cosine similarity as the metric. Similarly to other works \cite{simclr, vinyals, supcon}, we apply L2 normalization to the output of the neural network which in combination with the dot product used in the losses described below is equivalent to cosine similarity. Thus, the cosine similarity between two descriptors can be directly computed as the dot product. Moreover, this normalization helps contrastive losses to perform intrinsic hard positive/negative mining \cite{supcon}.

\begin{figure}
    \centering
    \includegraphics[width=0.99\textwidth]{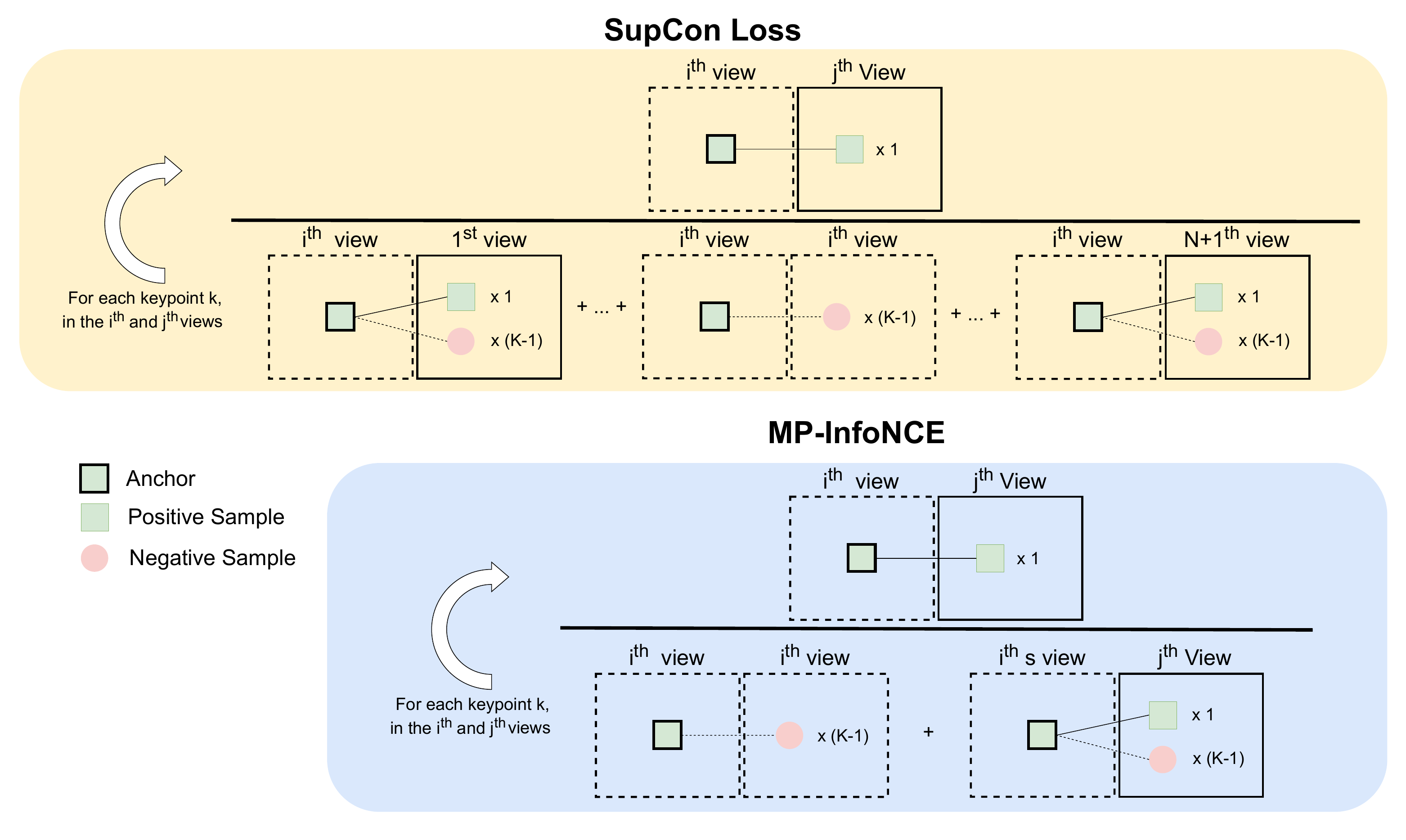}
    \caption{Schematic comparison between SupCon Loss and MP-InfoNCE, considering N+1 views, where N is the number of augmented versions of the original (1) image, and K keypoints per view, each of which matches with one keypoint on each other view (i.e. N positives per keypoint/sample).}
    \label{fig:schm}
\end{figure}

As training loss for ConKeD we propose two different alternatives: SupCon Loss \cite{supcon} and MP-InfoNCE \cite{vinyals,moco,simclr}. An schematic comparison between both losses can be seen in Figure \ref{fig:schm}.

\paragraph{SupCon Loss}
SupCon Loss \cite{supcon} is a loss function specifically designed for multi-positive multi-negative contrastive learning in the context of supervised image classification. This loss function can be directly applied to our proposal. For our framework, which uses a multiviewed batch of $N+1$ images, each of them containing $K$ keypoints, $k$ is the unique index of an arbitrary keypoint such that $k \in [1,K]$.
Therefore, the SupCon Loss function is defined as:

\begin{equation}
\mathcal{L}_{sup} =\frac{1}{N} \sum_{i = 1}^{N+1}{\sum\limits_{\substack{j = 1, \\ j \neq i}}^{N+1}}\sum_{k =1 }^{K}    -\log \frac{ exp(z_{ik} \cdot z_{jk} / \tau)}{\sum\limits_{\substack{c=1,\\c \neq k}}^{K}{exp(z_{ik} \cdot z_{ic}/\tau)} + \sum\limits_{\substack{l = 1, \\ l \neq i}}^{N+1}{\sum\limits_{c =1}^{K}{exp(z_{ik} \cdot z_{lc}/\tau)    }}},
\end{equation}

where we represent the network output for any keypoint as $z$. This way, $z_{ik}$, $z_{jk}$, $z_{lk}$  represent the output of network for keypoint $k$ in the views of index $i$, $j$ and $l$, respectively. We use the symbol $\cdot$ to denote the dot product. Finally, $\tau \in R$ is the scalar temperature parameter.

\paragraph{MP-InfoNCE Loss}

InfoNCE \cite{vinyals}, also known as NT-Xent, is a loss function commonly used in many metric learning methods \cite{vinyals,moco,simclr}. However, it is only suitable for the paradigm of a single positive and many negatives. Therefore we propose to modify it so that it can be applied in our framework. We define our version of InfoNCE, MP-InfoNCE as follows.

\begin{equation}
\mathcal{L}_{NCE} = \frac{1}{{N+1 \choose 2}K}\sum_{i =1}^{N+1} \sum_{\substack{j =1, \\ j>i}}^{N+1}\sum_{k = 1}^{K}{-\log \frac{exp(z_{ik} \cdot z_{jk}/ \tau)}{
\sum\limits_{\substack{c = 1,\\c \neq k}}^{K}{exp(z_{ik} \cdot z_{ic}/\tau)} + \sum\limits_{c = 1}^{K}{exp(z_{ik} \cdot z_{jc}/\tau)} }},
\end{equation}

where $N+1$ is the number of views in the our multiviewed batch ($1$ original image, $N$ augmented views). Since we compute the loss pairwise, the total number of combinations amounts to ${N+1 \choose 2}$. $i$ and $j$, such that $j>i$, are the indexes of two different images from the multiviewed batch forming a pair. $k$ represents the unique index of a keypoint such that $k \in [1,K]$. This way, $z_{ik}$ represents the output of network for keypoint $k$ in the view of index $i$. Likewise,  $z_{jk}$ represents the output of network for keypoint $k$ in the view of index $j$.

Overall, the main differences between SupCon Loss and MP-Info-NCE lie in the denominator of the innermost summation term. The denominator in SupCon Loss contains the matching similarities among all keypoints across all views, involving as many positives as augmented views N. Conversely, the denominator of MP-Info-NCE only contrasts against the matching similarities within the views involved in the numerator, and only involving one positive from one augmented view. Note that for N=1, i.e. 2 images, both losses would be equivalent.

\subsubsection{Keypoint matching and transformation computation}

After the keypoint extraction, the keypoint descriptors are computed by the descriptor network trained using ConKeD, which produces a dense block of descriptors for each image. Using the keypoints' locations, the descriptors for each keypoint are selected. Next, these descriptors are matched among the images in each registration pair. It should be noted that, for a descriptor to match another one, this match should be bidirectional. That is, a given descriptor A matches with another descriptor B, if both A is the closest to B and B is the closest to A. As in the training stage, the distance between descriptors is measured using cosine similarity. Additionally, as the keypoint detection network can accurately classify the keypoints in crossovers of bifurcations, we leverage this capability to ease and speed up the matching. In particular, we only match the descriptors of points within each class, that is, crossovers with crossovers and bifurcations with bifurcations. This drastically reduces the number of required computations, which was already low due to the high specificity of these keypoints.

After the descriptor matching step, the paired keypoints are used to estimate the geometric transformation using RANSAC \cite{ransac}. This is a well-known algorithm that is commonly used to estimate the parameters of mathematical models when there may be outliers in the data. In this case, the model is the transformation matrix between the images and the data are the matched keypoints. We employ a projective (or homographical) transformation, as previous works in the state-of-the-art \cite{rivas2, eccv20}.

\subsection{Experimental setup}

\subsubsection{Dataset}

For the training phase we employ the public DRIVE dataset. The images were obtained from a diabetic retinopathy screening program using a Canon CR5 non-mydriatic 3CCD camera with a FOV of 45 degrees. Each image has a resolution of $584\times565$ pixels. This dataset of 40 images is equally divided in the training and test sets, each of 20 images. This dataset has a ground truth of vessel crossovers and bifurcations that we use to train our keypoint detector \cite{Abbasi}. The original binary labeling from DRIVE is converted to heatmaps, as explained in \cite{alvaro_cmpb}.

For the testing phase we use FIRE, which is the only public CF dataset which has ground truth for registration. This ground truth consists of a set of labeled control points which can be used to compute the distance between the optimal and the produced registration. This dataset provides 134 image pairs, which can be divided in three categories. Category S has high overlapping between the images that compose each pair. On the contrary, category P has low overlapping. Finally, category A has high overlapping but the images show pathology progression within each pair. This complicates the registration but makes this category the most relevant one for clinical practice, where the registration is usually needed to monitor diseases. The total 134 image pairs are divided in 71 from category S, 49 from P and 14 from A.
Finally, it should be noted that there is an error in the ground truth provided by FIRE. In particular, a single point in an image in category P is incorrectly tagged. We opted to simply discard that point and evaluate the image with one less control point.

\subsubsection{Keypoint detection}

We train the keypoint detector network as in \cite{alvaro_cmpb}. That is,  we use the training set of the DRIVE dataset (i.e. 20 images), while reserving 25\% of the images for the validation set. The remaining 20 images compose the test set.
For this task, we use the U-Net \cite{ronneberger15} network architecture as in \cite{alvaro_cmpb}. To train the network, we use Adam \cite{adam} with learning rate decay. The network was trained from scratch. The learning rate is originally set to $1e-4$ with a patience of 2500 batches before reducing it.
Each batch contains a single image, and each time the learning rate is reduced, it is by a factor of 0.1. The training stops after the learning rate reaches  $1e-7$. 
We use spatial augmentation consisting of random affine transformations with the following parameters: random rotations of $\pm 90^{\circ}$, random scaling between $0.9-1.1 \times imageSize$ and random shearing of $\pm 20^{\circ}$. Finally, color augmentation is
also used randomly changing image components in the HSV color
space \cite{Liskowski}. 

The value for the intensity threshold was selected by testing multiple values and finding the one that provided the best F1-Score in the test set of the DRIVE dataset. In this case, 0.35 was found to be the optimal value \cite{alvaro_cmpb}.

\subsubsection{Keypoint description}

For the training, we use the whole DRIVE dataset (i.e. all the 40 images) in combination with the keypoint output produced by the detection network. Therefore, this network is trained using the DRIVE dataset resolution. As network architecture, we use the modified L2-net \cite{l2net} proposed in R2D2 \cite{r2d2}. This network produces a descriptor per input pixel without the need to upscale the descriptor block. Moreover, it has demonstrated to produce accurate descriptors in other works \cite{r2d2}. We train the network from scratch for 15000 epochs using Adam \cite{adam} with a constant learning rate of $1e-4$. For both losses, SupCon Loss and MP-InfoNCE, we set the temperature parameter $\tau =  0.1$. As augmentations we use random affine transformations with rotations of $\pm 60^{\circ}$, translations of $0.25\times imageSize$ in each axis, scaling between $0.75-1.25 \times imageSize$ and shearing of $\pm 30^{\circ}$. Moreover, we also include color augmentation in the form of random changes in the HSV color space \cite{hsv} as well as random Gaussian Noise with a mean of 0 and a standard deviation of 0.05. The geometric and color augmentations are applied to every augmented image while the random Gaussian noise is applied with a probability of 0.25. During training, the detection of keypoints is performed only on the original images. Therefore, the geometric augmentations are applied to both images and keypoints. Finally, regarding the multiviewed batch sizes, we perform experiments with 3 different sizes. Using a single original image, we try the multiviewed batch sizes 1+1 (i.e. 1 original image and 1 augmentation), 1+9, and 1+19. Moreover, we also train a descriptor network using the triplet loss, to use as baseline. In order for the comparison to be completely fair we use cosine similarity as the distance metric. Moreover, we use a margin of value 0.05, which we empirically found to be the most adequate. The training is done using the exact same settings described above except that, in this case, we use a learning rate of $1e-5$ and train the network during 75000 epochs.

\subsubsection{Keypoint matching}

After the description step, we match the descriptors for both images  composing a registration pair. The resulting keypoint matches are upscaled from the detection resolution (DRIVE) to the test resolution (FIRE). Once the keypoints are upscaled, they are used in combination with RANSAC \cite{ransac} to create a projective transformation that aligns the image pair. Therefore, the test images are aligned using their full size which is required to compare among methods. 

\subsection{Evaluation methodology}

To validate our approach, first we use the evaluation metric proposed by the authors of the FIRE dataset \cite{fire}. This metric, called Registration Score, measures the average Euclidean distance between the control points of each image pair after the registration process. 
Depending on whether the error of an image pair falls below a threshold, the pair is classified as either successful or unsuccessful. By plotting the ratio of successful registrations on the Y axis and progressively increasing the error threshold along the X axis (defined between 1 and 25 pixels for FIRE), we can create a 2D graph and compute its Area Under Curve (AUC), allowing for quick and easy comparison among methods \cite{rivas,rempe, votus, eccv20}.

We calculate this metric for each category separately as well as for the whole dataset as proposed in FIRE \cite{fire}. Moreover, in order to compare our results to \cite{eccv20} we also compute the average (Avg.) of the registration score among the three categories. We also propose to compute a weighted average (W. Avg.) among the three categories, using the number of images per category as the weighting factor. This metric should provide a more representative picture of the performance through the whole dataset.

The Registration Score only considers the error produced using all the detected keypoints. Thus, this metric may encourage brute-forcing both detection and description to improve the results, without considering the efficiency or computation time. Given an accurate enough description, simply detecting more points (i.e. higher keypoint frequency) improves the chance that each keypoint matches with another one in its approximate corresponding space, even if the detector is pseudo-random. However, this entails a much higher computational cost. Performing the registration with a smaller number of keypoints (i.e. more spaced out) puts more emphasis on both steps, as the detection needs to be repeatable enough so that the keypoints are in approximately the same locations in both image of a pair.

In order to perform a more complete evaluation, we propose a new metric based on the registration score that takes into account the number of matching keypoints. By limiting the amount of paired keypoints (i.e. after descriptor matching) used in RANSAC we can better evaluate the quality of the detected keypoints and their descriptors.
For this metric we propose to use the top-N keypoints, as ordered by the distance of their descriptors. That is, we take the N keypoints whose descriptors are closest, i.e. which offer a higher quality match. By increasing the value of N (i.e. the number of keypoint pairs) and computing the registration score for each one of the values, we can plot another curve and calculate its AUC. This way, we can evaluate the performance of the network across different number of keypoints. It should be noted that we cannot have less keypoints than the required by the transformation model, therefore $N>4$. Moreover, as we match the keypoints independently per class (i.e. crossovers and bifurcations), we compute the top-N points for each class. 
In particular, we start the evaluation with the top-3 keypoints per class, which is a total of 6 keypoints (i.e. the best 3 crossovers and the best 3 bifurcations). Next, we increase the number to top-4, which is 8 points in total and so on, up to top-25 (with a total of 50 keypoints). In this evaluation we set the RANSAC budget according to the amount of keypoints such that all RANSAC possibilities are computed. It should be noted that, in some cases, the value of N might be bigger than the number of available or matched keypoints for each class (either crossovers or bifurcations). In this case no more points are added to that specific class. We name this metric Variable Top Keypoint Registration Score (VTKRS).
Similarly to the registration score, we also compute the VTKRS per category and overall for the whole dataset.

\section{Results and discussion}
This section is structured as follows: first, in subsection \ref{ssec:bs} we evaluate our method. This includes a comparison with a triplet baseline, as well as analyzing the effect of including multiple positives as well as the effect of increasing the number of positives, by increasing the multiviewed batch size. Next, in subsection \ref{ssec:lf} we compare both of the proposed loss functions using the best multiviewed batch size. Finally, in subsection \ref{ssec:sota}, we compare our approach with the different state of the art methods.

\subsection{Evaluation of the proposed approach} \label{ssec:bs}

\begin{table}[tb]
\centering
\resizebox{0.85\textwidth}{!}{
\begin{tabular}{lcccccc} 
\toprule
           & FIRE           & A             & P              & S              & Avg.           & W. Avg.         \\ 
\hline
Triplet (SP-SN)         & 0.696          & 0.683         & 0.349          & 0.939          & 0.657          & 0.697           \\
Proposed 1+1 (SP-MN) & 0.748          & 0.743         & 0.469          & 0.942          & 0.718          & 0.748           \\
Proposed 1+9 (MP-MN)  & \textbf{0.755} & \textbf{0.76} & \textbf{0.477} & \textbf{0.946} & \textbf{0.728} & \textbf{0.755}  \\
Proposed 1+19 (MP-MN)  & 0.754          & 0.757         & \textbf{0.477} & 0.944          & 0.726          & 0.753
\\
\bottomrule
\end{tabular}}
\caption{Results measured in Registration Score AUC for the different number of image views sizes using SupCon Loss as well as the baseline with triplet loss. Best results in bold. SP means single-positive, SN means single-negative, MP means multi-positive and MN means multi-negative.}
\label{tab:resb}
\end{table}

\begin{table}[tb]
\centering
\resizebox{0.85\textwidth}{!}{
\begin{tabular}{lcccccc} 
\toprule
                            & FIRE           & A             & P              & S              & Avg.           & W. Avg.         \\ 
\hline
Triplet (SP-SN)           & 0.602          & 0.613         & 0.231           & 0.857          &  0.567              &  0.603         \\
Proposed 1+1 (SP-MN)   & 0.626          & 0.676         & 0.324           & 0.824          &  0.623        &    0.642        \\
Proposed 1+9 (MP-MN)   & \textbf{0.656} & 0.685           & \textbf{0.368}  & 0.849 & \textbf{0.634 } & \textbf{0.656}  \\
Proposed 1+19 (MP-MN) & 0.654          & \textbf{0.696}   & 0.344         & \textbf{0.862}         & 0.631          & 0.650           \\
\bottomrule
\end{tabular}}
\caption{Results measured in VTKRS for the different number of image views using SupCon Loss as well as the baseline with triplet loss. Best results in bold. SP means single-positive, SN means single-negative, MP means multi-positive and MN means multi-negative.}
\label{tab:resbvt}
\end{table}

In order to evaluate the performance of ConKeD, we compare our approach to the Triplet Loss, used in state-of-the art approaches \cite{eccv20}. Additionally, we compare three different variants of our proposal, the first one using a batch of 1+1 views such that our method is multi-negative and single-positive and the other two using batches of 1+9 and 1+19 views each such that our method is both multi-negative and multi-positive.
Therefore, we are comparing a single-negative single-positive (triplet) against a multi-negative single-positive (1+1 views) to the novel multi-positive multi-negative methods (1+9, 1+19). For this comparison we use SupCon Loss \cite{supcon}. Table \ref{tab:resb} shows the results for this experiment. 

Firstly, the results show that our proposal clearly outperforms the triplet loss approach, regardless of the multiviewed batch size. Moreover, we have empirically found that training with multi-positive multi-negative losses (either SupCon Loss or MP-InfoNCE) drastically reduces the training time, requiring five times less iterations to converge than the triplet loss.  In terms of categories in the FIRE dataset, the difference between methods is less notorious in the category S, which is more forgiving with inaccurate descriptor matches due to the high amount of overlapping. On the other hand, the difference in categories A and P is notable due to the disease progression and low overlapping. These characteristics make accurate descriptor matching more relevant, due to the lower amount of available keypoints. In the category P, there are less keypoints in the overlapping zone and, therefore, less usable keypoints in the registration. Thus, it is more important to correctly match the few usable keypoints, in order to be able to produce an accurate alignment. In category A, due to the progression of pathologies, the amount of keypoints is also relevant, as some may be vanished or obstructed. However, given the higher amount of overlapping, the number of keypoints is less limiting, putting more weight on the robust description that allows to recognize keypoints under different conditions (i.e. pathological progression). Therefore, to evaluate the descriptors, category A is the most relevant one. It should be noted that, as the detected keypoints are exactly the same in all the approaches, the differences are purely due to the improved descriptors.

Regarding the different multiviewed batch sizes, overall, the results are accurate and the differences between the different number of views are small but highly relevant. These differences in the results are constant across all the FIRE categories, creating a clear trend. In particular, the most notable difference is between the batch size 1+1 and the other two (1+9 and 1+19). In this case, the results on FIRE are around 1\% less with the 1+1 batch size than with bigger batch sizes. Similarly, in category P the difference among the 1+1 batch and the rest is around 1\% as well. For category A, the difference increases to almost 2\% while on category S, it is less than 1\%. This exemplifies the different level of difficulty in learning adequate descriptors for each category, as previously stated. Category S, having the most overlapping is the easiest, while A and P are more complex due to the pathologies and low overlapping. 

\begin{figure}[t]
    \centering
    \includegraphics[width=0.85\textwidth]{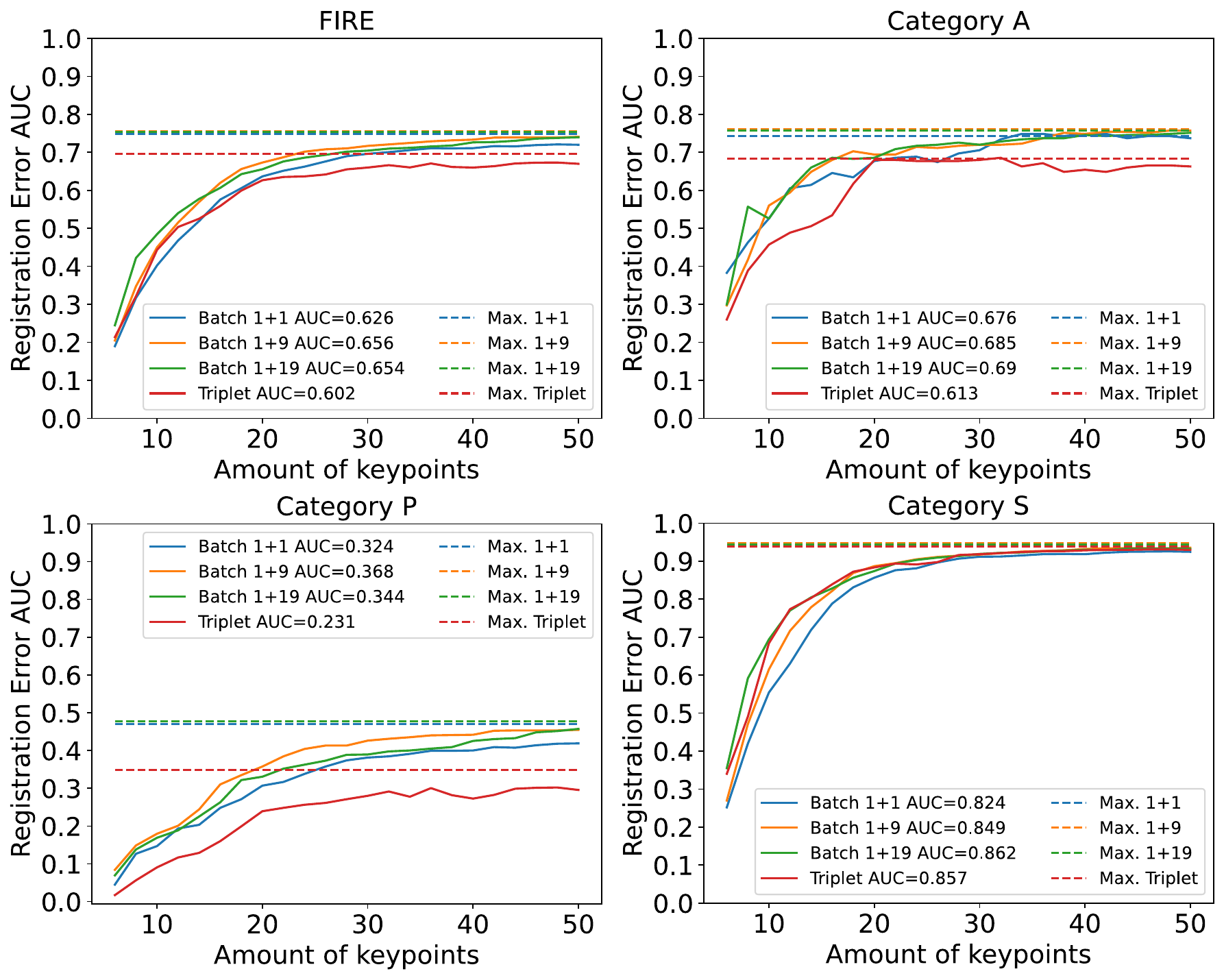}
    \caption{AUC curves for the different losses measured in VTKRS. In the legend, "Max." represents the maximum Registration Score AUC produced using all the keypoints.}
    \label{fig:vt}
\end{figure}

Table \ref{tab:resbvt} shows the results of our evaluation using VTKRS. These results are shown graphically in Fig. \ref{fig:vt}. Similarly to the previous analyses, the difference between our proposal and the triplet loss baseline is clear. In category S, where the descriptor matching is less relevant, the results for all the approaches are equivalent. However, in categories A and P, the differences are notable and can be clearly seen in Fig. \ref{fig:vt}, where the triplet VTKRS AUC is significantly lower than in our proposals. Overall, in FIRE, due to the prevalence of the category S, the difference between our method and the triplet baseline is less notable but still highly significant. Regarding the different variants of our proposal, it can be seen that adding multiple positives (i.e. 1+9 and 1+19) consistently improves the performance across all categories and, specially, in the more complicated ones (A and P).

Regarding the number of views in a multi-positive setting, (i.e. 1+9 and 1+19) the results are equivalent with both metrics. Moreover, given the extended training in our experimentation, i.e. 15000 epochs, we can ensure that the difference in performance between 1+1 and other two approaches is not due to the different number of seen image views. All three approaches are trained well beyond their convergence point, such that, there are no more meaningful improvements possible. This serves to isolate the improvements of the multi-positive approach from the potential improvements of using a higher number of views per batch and thus allowing the network to see more images per iteration. Therefore, the difference between adding multiple positives (1+1 vs the rest) and just increasing the number of views in the batch (1+9 vs 1+19) is clear. Given that the results of batches 1+9 and 1+19 are equivalent, we choose 1+9 as the reference batch to compare losses and models in the following sections.

\subsection{Comparison of loss functions}\label{ssec:lf}

\begin{figure}[t]
    \centering
    \includegraphics[width=0.9\textwidth]{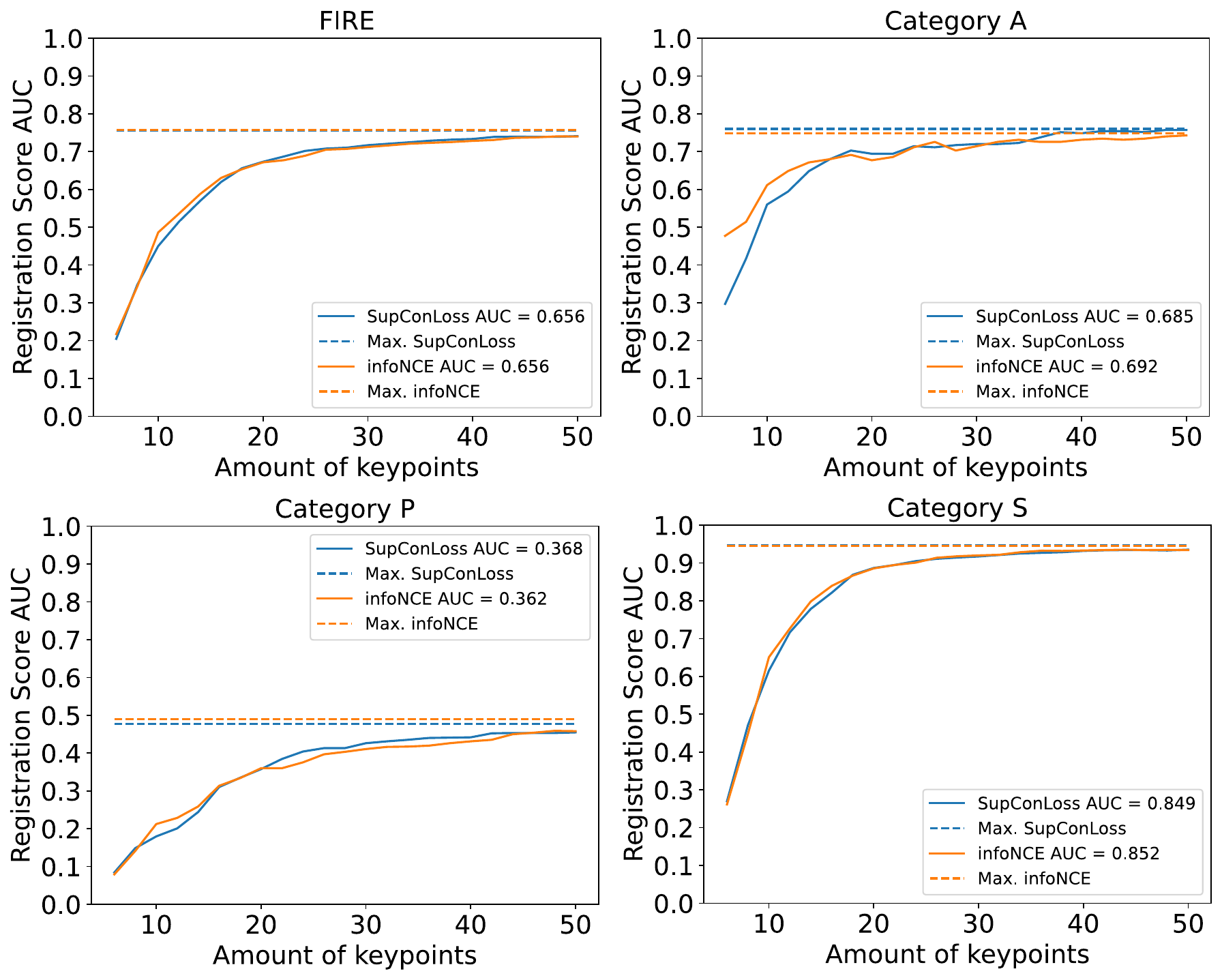}
    \caption{AUC curves for the different losses measured in VTKRS. In the legend, "Max." represents the maximum Registration Score AUC produced using all the keypoints.}
    \label{fig:aucs}
\end{figure}

In this section, we compare SupCon Loss against MP-InfoNCE. The experiments are performed using the optimal multiviewed batch size found in the previous experiments (i.e. 1+9). The results are detailed in Tables \ref{tab:resL} and \ref{tab:resL2} as well as in Fig. \ref{fig:aucs}. Representative examples of registered images from the FIRE dataset using our both losses can be seen in the top rows of Figure \ref{fig:exc}.

In terms of registration score over the FIRE dataset, detailed in Table  \ref{tab:resL}, the results for both losses are very similar. The biggest differences appear in Category A ($+1.1\%$ for SupCon) and Category P ($+1.2\%$ for MP-InfoNCE). However, these differences are small due to the limited nature of the registration score evaluation over FIRE.
The results in terms of VTKRS can be seen in Table \ref{tab:resL2} and in Fig. \ref{fig:aucs}. The final numeric differences in the AUC values for both losses are small, like in the previous evaluation. However, looking at the curves, the differences are much more evident. In particular, for category A, we can see that MP-InfoNCE provides results that are more accurate with a low number of keypoints. In category P both losses perform similarly, MP-InfoNCE obtains better results with less points while SupCon Loss improves more given more points. However, in the end, the two approaches become virtually even when many or all of the keypoints are used. 

The difference in performance in the Category A (the only one where the difference is notable) is especially relevant as this is the category with the most clinical relevance. RIR is commonly used for disease monitoring and study follow-up, therefore, any automated method should desirably be robust to pathological lesions. Additionally, Category A can be viewed as the hardest category to align due to the number and variety of lesions and diseases that severely alter the morphology of the ocular fundus. While Category P obtains the lowest scores, this is due to the lower overlapping and thus the lower amount of common keypoints on both images of the registration pair. However, in category A the detected keypoints can disappear or be altered by pathology progression. Therefore, improving the results in this category is desirable and even more with a low number of keypoints as this makes the network more robust to pathology progression which can completely occlude the keypoints. Given this benefit of MP-InfoNCE over SupCon Loss we will use the MP-InfoNCE as the method to compare our approach against the state of the art. Nevertheless, while there are differences between MP-InfoNCE and SupCon Loss, both loss terms produce satisfactory performance. Therefore, we can conclude that our multi-negative multi-positive approach is robust to the specific loss term used.

\begin{table}[t]
\centering
\begin{tabular}{lllllll} 
\toprule
            & FIRE           & A             & P              & S              & Avg.           & W. Avg.         \\ 
\hline
SupCon Loss & 0.755          & \textbf{0.76} & 0.477          & \textbf{0.946} & \textbf{0.728} & 0.755  \\
MP-InfoNCE     & \textbf{0.758} & 0.749         & \textbf{0.489} & 0.945          & \textbf{0.728} & \textbf{0.758}           \\
\bottomrule
\end{tabular}
\caption{Results for the different losses measured in Registration Score. Best results highlighted in bold.}
\label{tab:resL}
\end{table}

\begin{table}[t]
\centering

\begin{tabular}{lllllll} 
\toprule
            & FIRE           & A              & P              & S              & Avg.           & W. Avg.         \\ 
\hline
SupCon Loss & \textbf{0.656} & 0.685          & \textbf{0.368}          &0.849  & 0.634          & \textbf{0.656}           \\
MP-InfoNCE     & \textbf{0.656} & \textbf{0.692} & 0.362 & \textbf{0.852}           & \textbf{0.635} & \textbf{0.656}  \\
\bottomrule
\end{tabular}
\caption{Results for the different losses measured in VTKRS. Best results highlighted in bold.}\label{tab:resL2}
\end{table}

\subsection{Comparison with state-of-the-art registration methods}\label{ssec:sota}

\begin{figure}
    \centering
    \includegraphics[width=\textwidth]{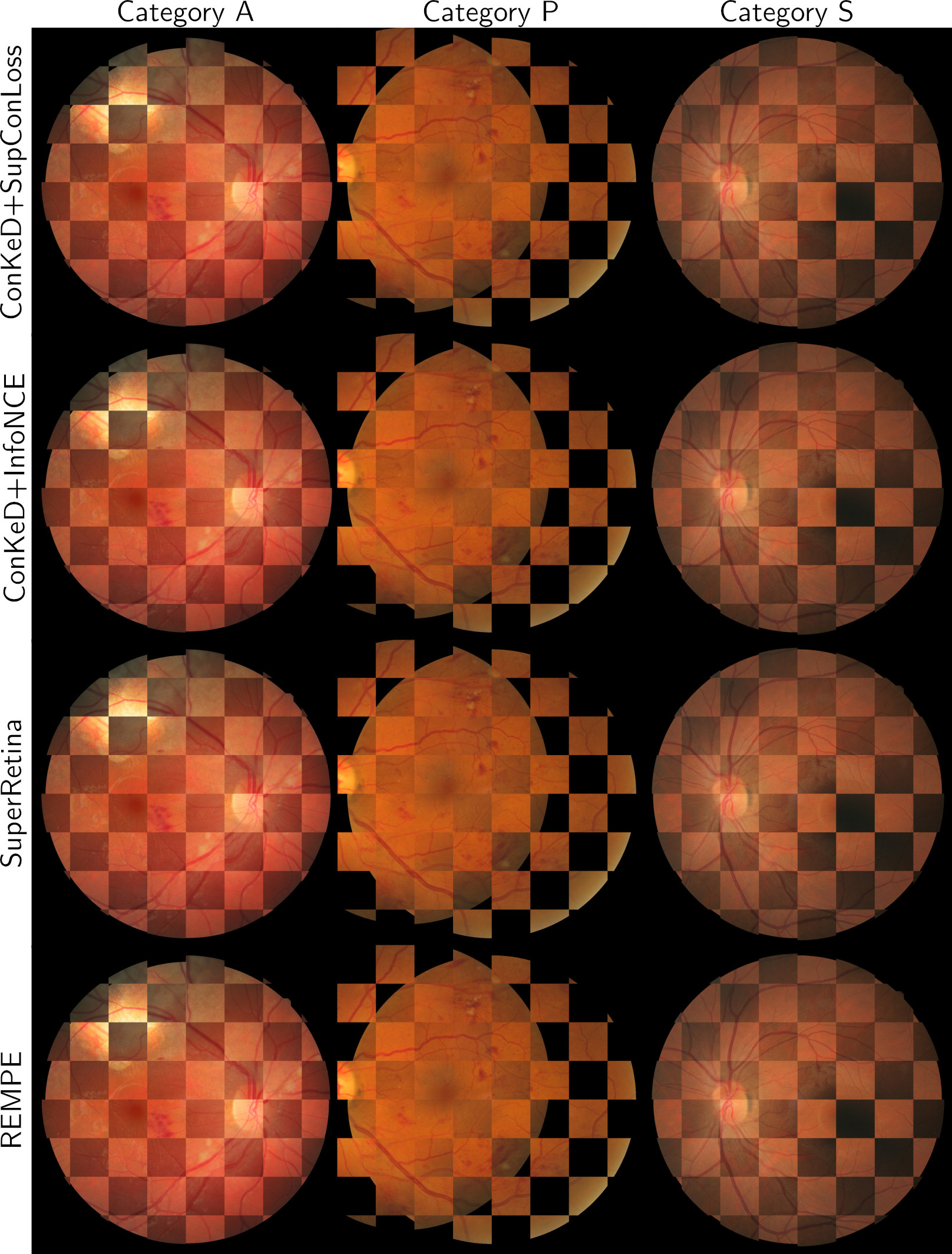}
    \caption{Representative 
    registration examples from the FIRE dataset for our proposals and different state-of-the-art methods.}
    \label{fig:exc}
\end{figure}

\begin{figure*}[!h]

\end{figure*}

\begin{table}[htb]
\centering
\resizebox{\textwidth}{!}{
\begin{tabular}{lcccccccc}
\hline
 & FIRE & A & P & S & Avg. & W. Avg. & Keypoints & \begin{tabular}[c]{@{}c@{}}Training \\ images\end{tabular} \\ \hline
VOTUS \cite{votus} & \textbf{0.812} & 0.681 & \textbf{0.672} & 0.934 & \textbf{0.762} & \textbf{0.811} & N/A & N/A \\
SR Manual \cite{eccv20} & - & 0.783 & 0.542* & 0.94 & 0.755* & 0.780* & 740 & 97 + 844 \\
SR PBO \cite{eccv20} & - & \textbf{0.789} & 0.516* & 0.944 & 0.750* & 0.773* & 740 & 97 + 844 \\
\textit{Proposed} & \textit{0.758} & \textit{0.749} & \textit{0.489} & \textit{0.945} & \textit{0.728} & 0.758 & 115 & 15 \& 40 \\
REMPE \cite{rempe} & 0.773 & 0.66 & 0.542 & \textbf{0.958} & 0.72 & 0.774 & - & N/A \\
Retina-R2D2 \cite{rivas2} & 0.695 & 0.726 & 0.352 & 0.925 & 0.645 & 0.575 & 5000 & 1748 \\
Rivas-Villar \cite{rivas} & 0.657 & 0.660 & 0.293 & 0.908 & 0.620 & 0.552 & 115 & 20 \\ \hline
\end{tabular}%
}
\caption{Comparison between our approach and state of the art methods, sorted by average. * indicates that the results were calculated using one image less. N/A indicates that the method does not use keypoints or does not require training samples. - indicates that the method is not publicly available and does not specify the number of keypoints used. $k+d$ denotes that k images are used for the detector training and $k+d$ images are used for the descriptor training. Meanwhile, $k \& d$ denotes that $k$ images are used for the detector training and $d$ images are used for the descriptor training. Best results in bold. Results sourced from the referenced papers.}
\label{tab:sota}
\end{table}

The comparison with the state-of-the-art is depicted in Table \ref{tab:sota}. In this comparison, we include the three best-performing approaches on the FIRE dataset as well as previous approaches using deep learning. 
The results show that the best performing method overall is VOTUS \cite{votus} and the best deep learning method is SuperRetina \cite{eccv20}. However, it should be noted that SuperRetina does not provide the AUC for the whole FIRE dataset as it is done in every other work in the state of the art \cite{votus, rempe, rivas}. Analyzing each category separately, we can see that in A our proposal obtains the second best results only behind SuperRetina by around 4\%. In category P both our method and SuperRetina produce significantly worse results than VOTUS \cite{votus}, a classical method that uses a quadratic transformation to help improving the alignment in these low overlapping cases. Finally, in category S, which is the easiest to register, the results of most methods are within a margin of around 1\%, therefore the differences are not notable. 

In terms of Registration Score, our method is the second best deep learning approach and overall provides results that are close to those reported by SuperRetina. Regarding the different categories, the most relevant difference is in category P where our method obtains 2.7\% less and 5.3\% less AUC than both SuperRetina methods. As explained, this is due to the lower number of keypoints detected per image. In the other categories, where the amount of keypoints is less relevant, our results are even closer, even surpassing SuperRetina on in category S. Importantly, our method provides several advantages and there are also some important experimental differences between both works that should be considered. SuperRetina trains its detector using a private dataset containing 97 images whereas we train the detector with just 15. Additionally, SuperRetina uses an extra set of 844 images to train the descriptor, for a total of 941, whereas we use just 40 images for our descriptor. Furthermore, SuperRetina requires extensive ad-hoc image pre-processing. Moreover, the descriptor computed by SuperRetina has 256 channels compared to 128 in our descriptor. Additionally, the results of SuperRetina depend greatly on the number of detected keypoints. As shown on Table \ref{tab:sota}, SuperRetina detects more than $6\times$ more keypoints than our approach. In that regard, the authors of SuperRetina also tested a variant of their method that detects approximately the same amount as our work (using just the labeled keypoints in their training set), obtaining a notable performance drop, from 0.755 to 0.685 (Avg.) in which case our method, which produces approximately the same number of keypoints, would obtain better results. Using more keypoints usually increases performance as it provides more chances for the descriptors to match and more paired keypoints to compute a transformation, at the cost of more computation. This is especially notable in category P where the small overlapping limits the number of keypoints usable for registration. Additionally, SuperRetina requires double inference (i.e. double run), including double detection, description, descriptor matching and transformation calculation. In contrast, our method does not require any of these extra steps making it more efficient.

Our method was designed with efficiency in mind as our multi-positive multi-negative training approach allows us to leverage more information during training, allowing for increased data efficiency. Additionally, through the use of domain-specific keypoints we can limit the cost of matching and transformation estimation. Overall, our proposed approach demonstrates execution times that are suitable for use in clinical day-to-day practice, providing fast image registrations. In particular, the average execution time for our whole methodology, including each step in the process is, approximately, 0.088 seconds per image pair. This time can be itself divided into the different steps of the method: 0.03 seconds for keypoint detection inference, 0.01 seconds for descriptor inference, 0.0006 for descriptor matching and 0.047 for transformation computation through RANSAC. In relation to this, our method detects, on average, 115 keypoints throughout the FIRE dataset. Given that the complexity of the descriptor matching stage is $O(n^2)$, this means that $115^2 = 13225$ comparisons are performed for each image pair. However, if our approach required the same amount of keypoints as SuperRetina (740 keypoints) or Retina-R2D2 (5000 keypoints), the number of comparisons required for the matching process would increase to 547,600 or 25,000,000, respectively. This would increase the matching time by factors of 41x and 1890x. Moreover, SuperRetina uses heavier descriptors and double inference, which would increase execution time and computational cost further. Our experiments revealed that running our method a second time does not improve the results. This evidences that our descriptors and keypoints are more robust to transformations since they are correctly calculated in a single pass, despite the larger transformations.

Finally, representative examples of registered images from the FIRE dataset using our approach as well as available state-of-the-art methods can be seen in Figure \ref{fig:exc}. While there are slight numerical differences between our approach and others, such as SuperRetina, they can hardly be visually spotted in the images. This is due to the high resolution of the images in FIRE. Small registration errors (i.e. 1 or 2 pixels) become really hard to spot on images of such size while significantly impacting numerical results.

In conclusion, overall, our method, with the novel ConKeD description framework, provides very similar results to SuperRetina while requiring significantly less training samples, avoiding the need for ad-hoc pre-processing, using far less keypoints and just requiring a single execution.

\section{Conclusion}

In this work we proposed ConKeD, the first multi-positive multi-negative contrastive descriptor learning approach for keypoint-based image registration. As it uses multiple positives and negatives, it can leverage additional information from the input images, which improves the resulting descriptors. We propose a complete methodology for the registration of CF images, using vessel crossovers and bifurcations as keypoints together with the descriptors learned using ConKeD. 

In order to validate our method, we perform several experiments in the well-known FIRE dataset and analyze several factors in the methodology. The results show that our proposal clearly outperforms triplet loss, the alternative typically used to learn descriptors in the state of the art.  In terms of the losses, we found that SupCon Loss and a multi-positive InfoNCE performed similarly with a slight advantage to the latter evidencing that our framework is robust to specific loss terms. Moreover, our method provides results competitive with the best deep learning state of the art method while using less training samples, less keypoints and not requiring double inference nor pre-processing.

As future work, it would be interesting to test more losses in combination with our ConKeD framework. In this regard, developing specific losses for the multi-positive paradigm is also desirable to try and improve the results. Additionally, a limitation of our current approach and of others in the state of the art is the reliance on inherently limited domain-specific keypoints, such as crossovers and bifurcations. These keypoints are not distributed evenly across the retina and may be completely absent in some parts. This can reduce the overall registration performance since there are no points to accurately drive the transformation in every part of the fundus. The arbitrary nature of these domain-specific keypoints may also reduce the performance of these methods on patients with less crossovers and bifurcations visible in the eye fundus. Therefore,  future works could focus on developing methods to increase the number of keypoints while keeping  a high specificity. This  could potentially improve the results, especially in category P, which has the lowest overlapping. In that regard, it could even be convenient to study the inclusion of generic keypoints. These, despite their generally lower specificity, could help regularize the registration if there are zones of the image that lack  domain-specific keypoints. Therefore, the inclusion of generic keypoints in a measured way may be beneficial without a significant reduction in the efficiency of the method.

\section*{Declarations}

The authors have no relevant financial or non-financial interests to disclose.

\section*{Acknowledgments}

This work is supported by Ministerio de Ciencia e Innovación, Government of Spain, through the \mbox{PID2019-108435RB-I00}, \mbox{TED2021-131201B-I00}, and \mbox{PDC2022-133132-I00} 
research projects; Consellería  de  Cultura,  Educación e Universidade, Xunta de Galicia, through Grupos de Referencia Competitiva ref. \mbox{ED431C 2020/24}, predoctoral fellowship ref. \mbox{ED481A 2021/147} and the postdoctoral fellowship ref. \mbox{ED481B-2022-025}. Funding for open access charge: Universidade da Coruña/CISUG.

\bibliographystyle{unsrt}
\bibliography{references}  

\begin{thebibliography}{10}

\bibitem{survey}
Max~A. Viergever, J.B.~Antoine Maintz, Stefan Klein, Keelin Murphy, Marius Staring, and Josien~P.W. Pluim.
\newblock A survey of medical image registration – under review.
\newblock {\em Medical Image Analysis}, 33:140--144, 2016.
\newblock 20th anniversary of the Medical Image Analysis journal (MedIA).

\bibitem{book_mir}
Joseph Hajnal, Derek Hill, and D.J. Hawkes.
\newblock {\em Medical image registration}.
\newblock Biomedical engineering series. CRC Press, Boca Raton, FL, 01 2001.

\bibitem{Narasimha}
Harihar Narasimha-Iyer, Ali Can, Badrinath Roysam, Howard~L. Tanenbaum, and Anna Majerovics.
\newblock Integrated analysis of vascular and nonvascular changes from color retinal fundus image sequences.
\newblock {\em IEEE Transactions on Biomedical Engineering}, 54(8):1436--1445, 2007.

\bibitem{forrester2020eye}
John~V Forrester, Andrew~D Dick, Paul~G McMenamin, Fiona Roberts, and Eric Pearlman.
\newblock {\em The eye e-book: basic sciences in practice}.
\newblock Elsevier Health Sciences, 2020.

\bibitem{RIR}
Carlos Hernandez-Matas, Xenophon Zabulis, and Antonis~A. Argyros.
\newblock Retinal image registration as a tool for supporting clinical applications.
\newblock {\em Computer Methods and Programs in Biomedicine}, 199:105900, 2021.

\bibitem{costeffec}
Ra~Ho, Lina~D. Song, Jin~A. Choi, and Donghyun Jee.
\newblock The cost-effectiveness of systematic screening for age-related macular degeneration in south korea.
\newblock {\em PLOS ONE}, 13(10):1--14, 10 2018.

\bibitem{Kanski}
J.F. Salmon.
\newblock {\em Kanski's Clinical Ophthalmology: A Systematic Approach}.
\newblock Elsevier, 2020.

\bibitem{rivas2}
David Rivas-Villar, Álvaro S.~Hervella, José Rouco, and Jorge Novo.
\newblock Joint keypoint detection and description network for color fundus image registration.
\newblock {\em Quantitative Imaging in Medicine and Surgery}.

\bibitem{Rivas-Villar:23}
David Rivas-Villar, Alice~R. Motschi, Michael Pircher, Christoph~K. Hitzenberger, Markus Schranz, Philipp~K. Roberts, Ursula Schmidt-Erfurth, and Hrvoje Bogunovi\'{c}.
\newblock Automated inter-device 3d oct image registration using deep learning and retinal layer segmentation.
\newblock {\em Biomed. Opt. Express}, 14(7):3726--3747, Jul 2023.

\bibitem{Pluim}
Josien P.~W. Pluim, J.~B.~Antoine Maintz, and Max~A. Viergever.
\newblock Image registration by maximization of combined mutual information and gradient information.
\newblock In Scott~L. Delp, Anthony~M. DiGoia, and Branislav Jaramaz, editors, {\em Medical Image Computing and Computer-Assisted Intervention -- MICCAI 2000}, pages 452--461, Berlin, Heidelberg, 2000. Springer Berlin Heidelberg.

\bibitem{Balakrishnan}
G.~{Balakrishnan}, A.~{Zhao}, M.~R. {Sabuncu}, A.~V. {Dalca}, and J.~{Guttag}.
\newblock An unsupervised learning model for deformable medical image registration.
\newblock In {\em 2018 IEEE/CVF Conference on Computer Vision and Pattern Recognition}, pages 9252--9260, 2018.

\bibitem{cheng18}
Xi~Cheng, Li~Zhang, and Yefeng Zheng.
\newblock Deep similarity learning for multimodal medical images.
\newblock {\em Computer Methods in Biomechanics and Biomedical Engineering: Imaging \& Visualization}, 6(3):248--252, 2018.

\bibitem{Haskins}
Grant Haskins, Uwe Kruger, and Pingkun Yan.
\newblock Deep learning in medical image registration: a survey.
\newblock {\em Machine Vision and Applications}, 31(1):8, Jan 2020.

\bibitem{voxel}
Guha Balakrishnan, Amy Zhao, Mert~R. Sabuncu, John Guttag, and Adrian~V. Dalca.
\newblock Voxelmorph: A learning framework for deformable medical image registration.
\newblock {\em IEEE Transactions on Medical Imaging}, 38(8):1788--1800, 2019.

\bibitem{Benvenuto}
Giovana~A. Benvenuto, Marilaine Colnago, Maurício~A. Dias, Rogério~G. Negri, Erivaldo~A. Silva, and Wallace Casaca.
\newblock A fully unsupervised deep learning framework for non-rigid fundus image registration.
\newblock {\em Bioengineering}, 9(8), 2022.

\bibitem{alvaro}
Álvaro S.~Hervella, José Rouco, Jorge Novo, and Marcos Ortega.
\newblock Multimodal registration of retinal images using domain-specific landmarks and vessel enhancement.
\newblock {\em Procedia Computer Science}, 126:97--104, 2018.
\newblock Knowledge-Based and Intelligent Information \& Engineering Systems: Proceedings of the 22nd International Conference, KES-2018, Belgrade, Serbia.

\bibitem{Rv-v}
David Rivas-Villar, Alice~R. Motschi, Michael Pircher, Christoph~K. Hitzenberger, Markus Schranz, Philipp~K. Roberts, Ursula Schmidt-Erfurth, and Hrvoje Bogunovi\'{c}.
\newblock Automated inter-device 3d oct image registration using deep learning and retinal layer segmentation.
\newblock {\em Biomed. Opt. Express}, 14(7):3726--3747, Jul 2023.

\bibitem{dlir}
Bob~D. {de Vos}, Floris~F. Berendsen, Max~A. Viergever, Hessam Sokooti, Marius Staring, and Ivana Išgum.
\newblock A deep learning framework for unsupervised affine and deformable image registration.
\newblock {\em Medical Image Analysis}, 52:128--143, 2019.

\bibitem{Haonan}
Haonan Xiao, Xinzhi Teng, Chenyang Liu, Tian Li, Ge~Ren, Ruijie Yang, Dinggang Shen, and Jing Cai.
\newblock A review of deep learning-based three-dimensional medical image registration methods.
\newblock {\em Quantitative Imaging in Medicine and Surgery}, 11(12), 2021.

\bibitem{rempe}
C.~{Hernandez-Matas}, X.~{Zabulis}, and A.~A. {Argyros}.
\newblock Rempe: Registration of retinal images through eye modelling and pose estimation.
\newblock {\em IEEE Journal of Biomedical and Health Informatics}, 24(12):3362--3373, 2020.

\bibitem{votus}
D.~{Motta}, W.~{Casaca}, and A.~{Paiva}.
\newblock Vessel optimal transport for automated alignment of retinal fundus images.
\newblock {\em IEEE Transactions on Image Processing}, 28(12):6154--6168, 2019.

\bibitem{ransac}
M.~Fischler and R.~Bolles.
\newblock Random sample consensus: a paradigm for model fitting with applications to image analysis and automated cartography.
\newblock {\em Commun. ACM}, 24:381--395, 1981.

\bibitem{rivas}
David Rivas-Villar, Álvaro S.~Hervella, José Rouco, and Jorge Novo.
\newblock Color fundus image registration using a learning-based domain-specific landmark detection methodology.
\newblock {\em Computers in Biology and Medicine}, 140:105101, 2022.

\bibitem{eccv20}
Jiazhen Liu, Xirong Li, Qijie Wei, Jie Xu, and Dayong Ding.
\newblock Semi-supervised keypoint detector and descriptor for retinal image matching.
\newblock In Shai Avidan, Gabriel Brostow, Moustapha Ciss{\'e}, Giovanni~Maria Farinella, and Tal Hassner, editors, {\em Computer Vision -- ECCV 2022}, pages 593--609, Cham, 2022. Springer Nature Switzerland.

\bibitem{zou}
Beiji Zou, Zhiyou He, Rongchang Zhao, Chengzhang Zhu, Wangmin Liao, and Shuo Li.
\newblock Non-rigid retinal image registration using an unsupervised structure-driven regression network.
\newblock {\em Neurocomputing}, 404:14--25, 2020.

\bibitem{superpoint}
Daniel DeTone, Tomasz Malisiewicz, and Andrew Rabinovich.
\newblock Superpoint: Self-supervised interest point detection and description.
\newblock In {\em Proceedings of the IEEE Conference on Computer Vision and Pattern Recognition (CVPR) Workshops}, June 2018.

\bibitem{pbo}
Hannu Oinonen, Heikki Forsvik, Pekka Ruusuvuori, Olli Yli-Harja, Ville Voipio, and Heikki Huttunen.
\newblock Identity verification based on vessel matching from fundus images.
\newblock In {\em 2010 IEEE International Conference on Image Processing}, pages 4089--4092, 2010.

\bibitem{CLAHE}
Stephen~M. Pizer, E.~Philip Amburn, John~D. Austin, Robert Cromartie, Ari Geselowitz, Trey Greer, Bart {ter Haar Romeny}, John~B. Zimmerman, and Karel Zuiderveld.
\newblock Adaptive histogram equalization and its variations.
\newblock {\em Computer Vision, Graphics, and Image Processing}, 39(3):355--368, 1987.

\bibitem{r2d2}
Jerome Revaud, Cesar De~Souza, Martin Humenberger, and Philippe Weinzaepfel.
\newblock R2d2: Reliable and repeatable detector and descriptor.
\newblock In H.~Wallach, H.~Larochelle, A.~Beygelzimer, F.~d\textquotesingle Alch\'{e}-Buc, E.~Fox, and R.~Garnett, editors, {\em Advances in Neural Information Processing Systems}, volume~32. Curran Associates, Inc., 2019.

\bibitem{supcon}
Prannay Khosla, Piotr Teterwak, Chen Wang, Aaron Sarna, Yonglong Tian, Phillip Isola, Aaron Maschinot, Ce~Liu, and Dilip Krishnan.
\newblock Supervised contrastive learning.
\newblock In H.~Larochelle, M.~Ranzato, R.~Hadsell, M.F. Balcan, and H.~Lin, editors, {\em Advances in Neural Information Processing Systems}, volume~33, pages 18661--18673. Curran Associates, Inc., 2020.

\bibitem{moco}
Kaiming He, Haoqi Fan, Yuxin Wu, Saining Xie, and Ross Girshick.
\newblock Momentum contrast for unsupervised visual representation learning.
\newblock In {\em Proceedings of the IEEE/CVF Conference on Computer Vision and Pattern Recognition (CVPR)}, June 2020.

\bibitem{simclr}
Ting Chen, Simon Kornblith, Mohammad Norouzi, and Geoffrey Hinton.
\newblock A simple framework for contrastive learning of visual representations.
\newblock In Hal~Daumé III and Aarti Singh, editors, {\em Proceedings of the 37th International Conference on Machine Learning}, volume 119 of {\em Proceedings of Machine Learning Research}, pages 1597--1607. PMLR, 13--18 Jul 2020.

\bibitem{alvaro_cmpb}
Álvaro S.~Hervella, José Rouco, Jorge Novo, Manuel~G. Penedo, and Marcos Ortega.
\newblock Deep multi-instance heatmap regression for the detection of retinal vessel crossings and bifurcations in eye fundus images.
\newblock {\em Computer Methods and Programs in Biomedicine}, 186:105201, 2020.

\bibitem{vinyals}
A{\"{a}}ron van~den Oord, Yazhe Li, and Oriol Vinyals.
\newblock Representation learning with contrastive predictive coding.
\newblock {\em CoRR}, abs/1807.03748, 2018.

\bibitem{Abbasi}
Samaneh Abbasi-Sureshjani, Iris Smit-Ockeloen, Erik Bekkers, Behdad Dashtbozorg, and Bart ter~Haar Romeny.
\newblock Automatic detection of vascular bifurcations and crossings in retinal images using orientation scores.
\newblock In {\em 2016 IEEE 13th International Symposium on Biomedical Imaging (ISBI)}, pages 189--192, 2016.

\bibitem{ronneberger15}
Olaf Ronneberger, Philipp Fischer, and Thomas Brox.
\newblock U-net: Convolutional networks for biomedical image segmentation.
\newblock In Nassir Navab, Joachim Hornegger, William~M. Wells, and Alejandro~F. Frangi, editors, {\em Medical Image Computing and Computer-Assisted Intervention -- MICCAI 2015}, pages 234--241, Cham, 2015. Springer International Publishing.

\bibitem{adam}
Diederik Kingma and Jimmy Ba.
\newblock Adam: A method for stochastic optimization.
\newblock In {\em International Conference on Learning Representations (ICLR)}, 12 2015.

\bibitem{Liskowski}
Paweł Liskowski and Krzysztof Krawiec.
\newblock Segmenting retinal blood vessels with deep neural networks.
\newblock {\em IEEE Transactions on Medical Imaging}, 35(11):2369--2380, 2016.

\bibitem{l2net}
Yurun Tian, Bin Fan, and Fuchao Wu.
\newblock L2-net: Deep learning of discriminative patch descriptor in euclidean space.
\newblock In {\em 2017 IEEE Conference on Computer Vision and Pattern Recognition (CVPR)}, pages 6128--6136, 2017.

\bibitem{hsv}
Álvaro S.~Hervella, José Rouco, Jorge Novo, and Marcos Ortega.
\newblock Self-supervised multimodal reconstruction of retinal images over paired datasets.
\newblock {\em Expert Systems with Applications}, 161:113674, 2020.

\bibitem{fire}
Carlos Hernandez-Matas, Xenophon Zabulis, Areti Triantafyllou, Panagiota Anyfanti, Stella Douma, and Antonis Argyros.
\newblock Fire: Fundus image registration dataset.
\newblock {\em Journal for Modeling in Opthalmology (to appear)}, 01 2017.

\end{thebibliography}

\end{document}